# Implementing Human-like Intuition Mechanism in Artificial Intelligence

**Jitesh Dundas[1] & David Chik[2]**

[1]Edencore Technologies Ltd.
Row House 6, Opp Ambo Vihar, Tirupati Nagar-II, Virar (w), Thane-401303, India
Email: jbdundas@gmail.com

[2]Riken Institute
Wako-shi, Saitama, Japan

**Abstract**
Human intuition has been simulated by several research projects using artificial intelligence techniques. Most of these algorithms or models lack the ability to handle complications or diversions. Moreover, they also do not explain the factors influencing intuition and the accuracy of the results from this process. In this paper, we present a simple series based model for implementation of human-like intuition using the principles of connectivity and unknown entities. By using Poker hand datasets and Car evaluation datasets, we compare the performance of some well-known models with our intuition model. The aim of the experiment was to predict the maximum accurate answers using intuition based models. We found that the presence of unknown entities, diversion from the current problem scenario, and identifying weakness without the normal logic based execution, greatly affects the reliability of the answers. Generally, the intuition based models cannot be a substitute for the logic based mechanisms in handling such problems. The intuition can only act as a support for an ongoing logic based model that processes all the steps in a sequential manner. However, when time and computational cost are very strict constraints, this intuition based model becomes extremely important and useful, because it can give a reasonably good performance. Factors affecting intuition are analyzed and interpreted through our model.

## 1. Introduction

One of the serious problems in machine learning is the ability to understand and interpret past knowledge for accurately solving current problems or predicting possible events. Current algorithms and models cannot obtain the results as good as human intuition does. Most of these models are logic driven and are time dependent. They lack the ability to give consistently accurate results because on one hand, when information is not sufficient for drawing any conclusion, logic process simply gets stuck. On the other hand, time is a crucial constraint for real life scenarios, and logic process is slow because it has a large search space and a lot of calculation steps. These constraints indicate a serious need for faster models to resolve such limitations in machine learning.

Little work has been done in the study of the intuition based methods in AI and machine learning. Kahnemann explained the variations in statistical intuition and statistical knowledge [1]. His work also shows the possible mistakes in the intuition of people [1]. Clearly, intuition as a process is prone to incorrect values, and the correctness depends on various factors, especially the mapping of the correct element of the past experiences (or combination of them). Kahnemann cites the experiments done by Shane Frederick [1] which show that people have a tendency of not thinking hard. They seem to be inclined towards accepting what comes first in their mind without any proper and logical thinking.

Common Sense is defined as the ability to perceive possible consequences in a short period of time from a wide range of possibilities [2]. This explanation assures us that the normal process of thinking is based on the same principles. However, this explanation does not throw any light on the concept of intuition with a mathematical or logical explanation. Sloman [3] argues that the idea of comparing intuition with the concept of simulation, perception using analogical representation, is prone to several loopholes. These include issues in non-logical reasoning [3], which is a central issue in intuition. Sloman [3] further argues that philosophy cannot be related to AI for finding answers to such higher level intelligence e.g. intuition. Moreover, the concept of intuition as suggested by the above experts focus largely on the concept and not on the representation and use of entities in the process. The concept of unknown entities [4, 5] needs to be handled properly by them in order to be able to consider intuition as an effective problem-solving strategy. One of these examples is IBSEAD, which considers the presence of unknown entities for problem solving. Such concept of valid inference has not been considered for unknown entities by Sloman [3] in his views on AI and Philosophy.

Kolata [6] explains how the difference in thinking and approach between established AI pioneers created issues in the overall strategy to solve the AI problem – to make human like thinking machines. However, the paper did not throw any light on the possible reasons for the failure to map intuition till date. Minsky presented a deep work on frame based representation [7,8]. He explained the concept of thinking and thoughts as "frames" that keep changing with time [7]. Also, he explains how the entire concept of "falling in love" [8] is actually bypassing the mechanisms of selecting "the best and optimum choice" and making the "other unconfirmed choice" (maybe or may not be the case that this is the best choice) as the best choice. Minsky's approach to consider the cases of non-logical reasoning is quite interesting and effective in handling higher level human functions such as novelty, creativity and intuition. However Minsky does not explain the concept of unknown entities or the ability to reason for intuition and such higher level functions of human thinking. The concepts are quite interesting to read and understand, but they do not explain how to get them to implement intuition.

Herbert Simon was one of the first pioneers who actually came close to the concept of mapping of entities [9]. He explains the approach for handling intuition in the form of the novice user and the expert user. He further adds that knowledge and past experience are very important for intuition to be accurate. He described intuition as "subconscious pattern recognition" but failed to explain the concept in terms of how it mapped to the problems and the evolving nature of entities as well as that of the environment. He also fails to consider the scenarios where external environment and other related entities maybe affecting the intuition capacity of the user. Thus, Simon's explanation was quite pioneering but lacked the practical implementation to further his interesting work. The algorithm we propose in this paper is a step towards this direction.

The work done by Wang on Non-Axiomatic Logic [10] has explained logic to be made up of language, rules and semantics. His work again fails to explain the concept of intuition in terms of mathematical representation. McCarthy [11] explains how the concept of common sense can be explained using mathematical logic. He emphasizes the need for common sense logic rather than scientific theories. This supports the claim in this paper that the models and the theories that have been developed by the AI pioneers seem to fail in explaining the higher level functions such as intuition with a successful practical implementation.

Simon [12] explained how intuition is dependent on past knowledge and experience for better recall of solutions to the given problems or normal logical process. He [13] has defined intuition as a process that happens suddenly and which does not follow the normal steps of logical thinking. He also mentions about EPAM, a system that simulated the *human role verbal learning*. EPAM, developed in 1960s, used discrimination nets and improvisation based on the knowledge received from experience, to make better decisions. Simon asserts that intuition is just a form of recognition and that people with knowledge and experience in one field are able to use intuition better than novice people. However, this assertion does not explain the reason why sometimes common people are able to predict some of the most complicated problems in the world. Simon fails to look at the aspect where the experience of another field or area can be mapped by the human brain to a problem in another field. In a true experience, a person playing card bridge game for the first time ever, was able to win over a team of 14 people. This is not possible as per the explanation given by Simon [13] as the common person was able to do well right from the first attempt. Clearly, the person's past experience in **another** area must have helped him in developing his intuition better. Our intuition model proposes to consider such scenarios as well and give a more holistic approach. Moreover, the intuition model is based on mapping and pattern recognition [12, 13] while normal thinking processes involve a series of logically executable steps with reliable answers and proof to support the implemented logic.

McCormick [14] explains how the experts develop generalizations of six types to help in their specific method of getting better solutions to problems. However, this explanation lacks the details on the ability of intuition to get solutions to problems in areas of other topics, i.e. topics in which the user has no or less knowledge available. Sonntag [15] explains how an intuitive system with multimodal dialogue can be implemented by using intuition as a recommendation system. Sonntag believes that intuition can have a positive impact on the human like communication in which the user is continuously changing his behavior and dialogues based on attitude, experience and knowledge. The authors join Sonntag in this belief that intuition can play a very important role in the multimodal dialogues based interactive system in the future using the concept of intuition in them. Intuition is referred as being evolution based recognition [16] of patterns which consider past experience rather than logical thinking for their solutions. Although the role of neural nets is quite well accepted, it is still not clear as to how intuition occurs and the exact process that performs it. This paper is a step in this direction.

## 2. Methodology

Intuition offers the ability to obtain answers much faster than the normal process of logic-based thinking. Here, we propose an intuition based model which attempts to simulate intuition for quickly obtaining accurate results for a given dataset. We downloaded the datasets for Car Evaluation and Poker Game from the online UCI repository (http://archive.ics.uci.edu/ml/datasets). This model will be explained in Section 2.2 (Experimental Detail). First, let us explain the theory and formulation of this intuition model.

## 2.1 Theory and Formulation

We consider the following sets for explaining our model:-
1) Problem Set at time t ---> $A_t$ = {-∞,…, @, $, *, % , 4, 6 , 555, 0.333, -3.444, -4,…………………., ∞ }
2) Experience Set at time t---> $B_t$ = {-∞,… ,1, 2 ,S, R, L, 8, 9, #, 1.2 , -0.44 ,…………………………. ∞}

Please note that we consider the above sets as dynamic and changing values at time "t". The above elements and sets may be of any type, dimension, or value. We represent them in simple static elements as a simple representation of the corresponding problem. For example, the problem could be about "What is the expected score that this batting team will get in this match?" or "What is the expected GPA score that I will be able to get this semester". There are several such problems and questions that can exist in the real brain and we propose to represent them here as Problem Set elements for simplicity. Each of the problem set elements shows a representation of such a question or problem. Moreover, the vast knowledge in the brain is represented as Experience Set elements. Each of the experience is representing values as done in the problem set (the representation is similar to the problem set). For example, the number "4.0/5.0" could be one of the answers to the past question on "How much GPA score did I get last year?" or "How much GPA score did I get in my first year?" (Note that the question is **not** as same as the current problem. This is the past experience and shall store past values only. It does not guarantee the end solution, but maybe a means to the latter). All such values are mapped as elements in the Experience Set.

We also propose that one normal process may not be enough to execute one intuition process. Sometimes, there are multiple normal processes which may act together and dependent for a single intuition process. The work of the intuition model is to find the right solution using the correct experience element in the Experience set, process it as per its methodology and then send it as final solution to the current problem at hand.

Figure 1 explains the intuition model with a single problem set element at hand for ease of explanation. We believe that there are two processes that are carried out in the brain. One of them is the normal process (referred to as NP in below) and intuition based process (IP). The NP carries out the processing of problems using normal logic based approaches, in which the information related to the problem is gathered and connected to infer the solution. But the IP uses a different approach – a comparison based approach – which retrieves values from past experience that could serve as the solution for the given problem.

Current approaches of intuition have a very logical and analytical form of processing. However, in reality, intuition is known to be **symbolic and artistic** rather than logical. Intuition may take an "O" as a circle while a logical normal process takes this to be zero or an English alphabet. Intuition does not use the normal logic based approaches, but it uses basic mapping of the past experience in similar problem as an attempt to solve the current problem at hand. Therefore IP can be considered as a mapping function with the required adjustments and weight factors at the given time 't'.

Past experience is a major factor in shaping the intuition of the entity or person. The brain has to simply map it to the past experience and the result is sent back to the user after a minor modification to the answer. The brain continuously manages to learn process and store value from the past experience. It may also develop multiple interpretations for the same experience set after a period of time. This may be due to a change in the element's priority, importance or even mapping as the correct intuition result. For example, in the game of soccer, the brain has developed two ways of finding the possible winner of the match. One way is to use the normal process, where past statistics are seen and understood such as past games won or lost by each side, players and their strengths, etc. In the second method, the brain looks at the teams and just maps it to a past experience similar but not necessarily identical to the current problem (for example, the team wearing blue shirts looks better because of other experience not related to soccer). The mapping returns a value based on the confidence level, priority and importance level. Thus, the accuracy of the intuition depends on the correct choice of experience that fits the current scenario, confidence and importance to the brain of the problem. If these are correct, then the intuition gives correct value.

The steps in our intuition model (Figure 1) are as follows:-
1) Obtain an element from the problem set.
2) Obtain an element from the experience set based on a mapping.
3) Obtain the importance, priority of the processes to obtain the probabilistic value of the dependent thinking processes.
4) Obtain the secondary thought processes (intuition based or normal based process) into the considered formulae in the same manner. Note that these will be dependent processes.
5) Apply the adjustment factor on all the considered processes. Calculate the final answer.
6) Check if there are any external influences that change the values and then present the answer to the user after these final adjustments. This may include mental balance of the human or machine entity, thinking capacity(to undertake intuition), etc.

The formula to find the intuition based result of problem is given by:

$$f(x)_t = \text{Mapping Fn}( f(x)_t ) + \text{Adjustment Factor}$$

where,

$$\text{Mapping Fn} (f(x)_t) = [P (IP/NP) * \text{Importance} (IP) + \text{Priority (Exp. Set element)}] + [\text{Exp Set element value}] + P (\text{External Changes Factors})$$

where,

1) $f(x)_t$ = the functional representation of intuition at time 't'.
2) P (IP/NP) = Probability that IP happens in the presence of NP. We strongly believe that IP cannot exist or happen without the prior presence or execution of the NP process. Note that there can be multiple normal processes dependent for a single intuition process.
   ⇨ P (IP/NP) = P (IP / NP1 * NP2 * NP3….∞)
3) IP = intuition based process. It is the process that handles the intuition based model.
4) NP = normal based process. It is the process that handles the normal functioning of the brain. It involves normal calculations and logical thinking. The implementation of thinking process may use algorithms such as neural networks (NN), decision trees, bayesian inference, hidden markov models (HMM), etc. We have considered NN, HMM for our experiments in this paper.
   5) Priority (Exp Set element) = This term defines the priority of the experiment set element(experience knowledge set) that has been mapped to the problem set element as the solution.
   6) Imp (IP) = This variable defines the importance of the IP process to the given problem element set value.
   7) Exp ( or Experiment ) Set Element Value – The value of the element that best defines the past experience , matching the closest, with the problem set element currently presented.
   8) P (External Changes Factor) – Changes in the external factors that affect the final intuition process.

Please note that all the values in the above formulae are represented on a scale of (1-10) except for Actual Result, Desired Result and "t". Priority and Importance are two different variables in the above formulae. Consider 3 elements that identify the closest as the possible solution values for the given problem set. Which is the value that will be picked up? Here, we select the one with the highest priority (or closeness to the problem). Also, the Importance here is defined by the "importance of the element value to the topic". Thus, an element may have high priority to the given problem but its Intuition Process (IP) value could be low in importance to the problem.

The problem set explains the problem which is to be solved here. This is **the current problem at hand** with reference to the user. Let us consider the problem set element A "$" as the current problem. As shown in Figure 1, the IP now has to identify the knowledge set element value. Let us consider that "#" is the element to be considered here. Thus we put the values in the equation as:

$$f(x)_t = \text{Mapping Fn} (f(x)_t) + \text{Adjustment Factor}$$

Where,
$$\text{Mapping Fn} (f(x)_t) = [P (IP/NP) * \text{Importance} (IP) + \text{Priority (Exp. Set element)}] + [\text{Exp Set element value}] + P (\text{External Changes Factors})$$

Now**,** there is only one normal process mapped to the intuition process in this case as this is just a symbolic mapping. Thus, as an example, we can get:

1) P (IP/NP) = 7/10 (This is a probabilistic value and means that the probability that IP exists when there is NP already present)
2) Importance (IP) = 8/10.
3) Priority (Exp. Set element) = 7/10
4) Exp Set element value] = #
5) P (External Changes Factors) = 8/10

   ⇨ Mapping Fn $(f(x)_t)$ = [7/10 * 8/10] + 7/10 + # + 8/10
   ⇨ Mapping Fn $(f(x)_t)$ = 0.78 + 0.7 + # + 0.8
   ⇨ Mapping Fn $(f(x)_t)$ = 2.28 + #

Thus, we have the value as 2.28 + #. Note that the answer means that there is a change of 2.28 units in the mapped value of experience set element. If "#" was a numeric element, then 2.28 will be an addition to the mapped element answer. Thus, if # = 70, then the intuition model answer would be 70 + 2.28 = 72.28. If # was a symbol, then 2.28 will be a change in symbol value or attributes.

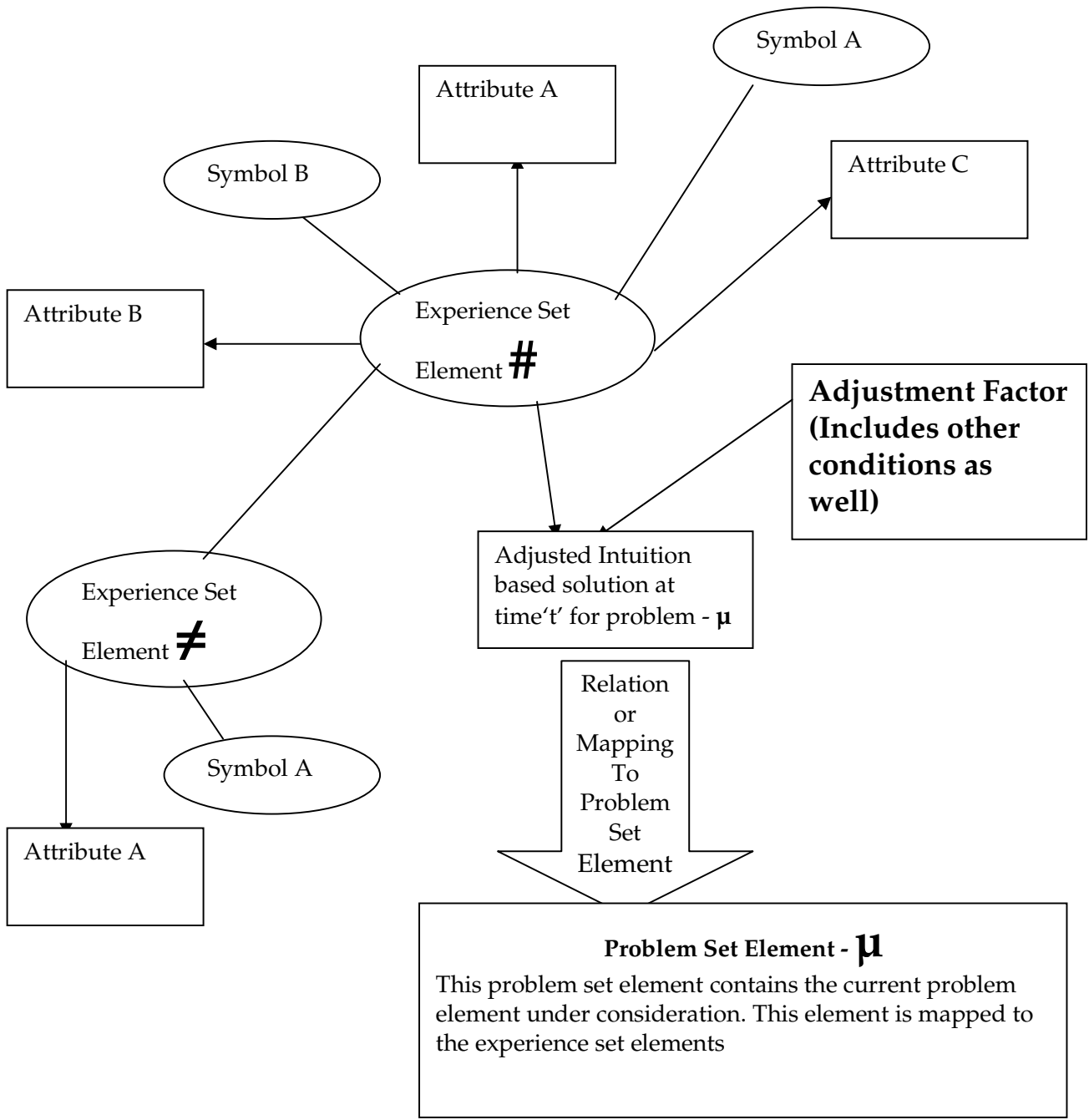

Figure 1. The intuition based process model. The space of intuition contains relational mappings between experience set elements and their associated attributes. The experience set element # is related to experience set element ≠. Note the values are taken and then adjusted against time and error handling to get the final value.

## 2.2 Experimental Detail

We use two datasets from the UCI repository:-
1) The car evaluation set: - Look at a car and then decide if it is good or not.
2) The poker hand set: - Look at the cards available and then estimate the winning hand combination.

Let us consider the poker hand set. This is extremely tricky and difficult for even humans to predict. For Poker datasets, we have to predict the possible hand that might happen. Now, intuition will help us relate the past experience set and the result thus might come positive. Thus, in each case we have a hypothesis and thus a positive answer is expected. Please note that we have unknown entities here. Before all the hands are in the hands of the user, we think of the user's possible third hand. But the third hand may or may not have been already dealt to the user yet. Thus, the third hand (or some of the poker cards) may be hidden (if already given to the user) or maybe unknown (not dealt yet and we do not know yet. What if suddenly you decide to change the card decks and opt for a new one for the rest of them? Uncertainty can also lead to unknown entities and not just hidden entities). After keeping into consideration these cases (put an equal number of cases that consider all these type of conditions) i.e. known, unknown and hidden entities, try to run the formulae on it, we expect an improvement. We have made the following assumptions:-

1) We called a neural network (NN) or a Hidden Markov model (HMM) as "untrained" when they assume the card set is fixed, so they try to calculate the probability based on a naive condition. In reality the card deck cannot be considered fixed, because the player does not know how many cards have been drawn before, and another player may suddenly request a new deck, etc. The calculation of probability of cards cannot be exact due to these uncertainties.

2) In the Intuition Process, there is a mapping function which fetches elements from the knowledge set to the experience set. We assume:-

If ((Importance (NP) > Importance (IP)) && Importance (IP) < 5)
    Then Experiment Set Element (as per the current NP) will give the wrong answer.

If ((Importance (IP) > Importance (NP)) && Importance (IP) > 5)
    Then Experiment Set Element (as per the current NP) will give the correct answer (i.e. intuition guesses it right).

If ((Importance (IP) > Importance (NP)) && Importance (IP) < 5)
    Then Experiment Set Element (as per the current NP) will give an answer adjusted to nearest possible value to the correct answer.

In other cases, we can get highly inaccurate answers. Also, the first version of the values is considered best as after that changes in attributes causes dilution of the Experiment set element value.

3) NN and HMM can be "trained" to include knowledge of other possibilities which are not considered before e.g. another player is more likely to change card deck when he loses a certain amount of money, etc. so that the performance of normal logical process can improve.

4) The values are obtained by performing a dry run on the considered datasets.

Regarding Car Evaluation dataset, there are also hidden and unknown entities. Unknown entities can include a wrong entry. For example, suppose the dataset is mixed with data about buses or trucks apart from cars. Normal process will either be too generalized or too specialized. If it is too generalized, the performance of evaluation will be low. If it is too specialized, it will fail saying that the data (bus) is unexpected or not valid. However, in case of intuition, the mapping will relate to the best fit for the bus or the truck. It somehow bypasses one filter (whether this is bus or car, but this classification is irrelevant to the goal of evaluation) and allow other filters (the essence of car evaluation) to operate properly. Thus, the accuracy can be expected higher in such cases as the number of attempts increase. In case of normal logical algorithms, if there is no past knowledge about trucks or buses, this will cause them to stop or give wrong answers.

The following is the method by which we have obtained our results:-

a) Poker Hand Datasets:-

The aim here is to predict the possible successful hand beforehand. Note that the steps below were performed 5 times (5 cycles) for each technique i.e. NN, HMM and Intuition Model.

1.1) The first card is issued and an attempt is made to find the possible hand using NN, HMM and Intuition Model.
1.2) The result is then recorded using the steps (2.1 - 2.4) which will be explained below.
1.3) The next card is then issued and the whole cycle (1.1 - 1.3) is repeated again.

For the untrained (naïve) conditions, the steps were as follows:-

2.1) We take the first record.
2.2) We then performed the NN technique to work on this naïve condition.
2.3) Then we try to predict the possible hand based on the current numbers of cards given.
2.4) The answer is then matched against the actual answer obtained at the end of the game (issue of the hand).

The above process ( 2.1 - 2.4) is repeated for HMM and intuition process also.

In the trained conditions, unknown possibilities as well as normal logical probability are considered and executed, and the same above steps are followed again.

b) Car Evaluation Dataset

The aim is to evaluate the car quality based on the details given. The details are given in the following format:-

3.1) The first car is taken into consideration.
3.2) First, a single quality information is given e.g. color.

3.3) Based on this information, the NN is performed and made to record the quality of the car.
3.4) The quality of the car is then recorded and then compared with the actual result (when all the qualities are available and the most accurate answer is available)
3.5) After this, the next quality is also made available to NN.
3.6) The NN is made to judge the quality of the car based on the two qualities. However, the result of the first iteration (when only one quality was available for the car) is not retained or allowed to influence in this case of untrained datatsets. This knowledge is retained in the case of trained datasets.
3.7) The result is recorded and then the third quality is made available in the same. The results are recorded again in the same way as in the steps 3.1 - 3.4.
3.8) The entire process in repeated in the same way.

The results were recorded and on the basis of the comparison of the percentage of mistakes or incorrect results. This is given by:-

Percentage of Errors = ( (No. of Mistakes) / (Actual Correct Answer) ) * 100

## 3. Results and Discussions

The results are summarized in Table 1. We found that the time was used less in getting the results in the proposed model as it was not doing logic based executions. Also, the accuracy was comparatively high but not as correct as the trained logic based methods. The intuition model had a higher level of accuracy in defending the solution of the given problem due to its ability to map non-logical solutions as optimum fits for the problem. There was no such ability in the current logical algorithms for such conditions. Hidden Markov Models (HMMs) were able to perform better due to their ability to consider hidden entities (entities that were already present but not considered in the solution/problem space). Note that these are untrained conditions with very less time for execution and unknown/partially unknown environments. In such cases, neural networks failed in these conditions by 30 -40%. However, HMMs failed by 20 – 30% while our intuition model failed by 10- 15 %. Also, our intuition model took less time of execution.

However, when time was not a constraint, the traditional models performed better than the intuition model. In trained conditions, as shown in Table 1, trained models gave an error of only 4-5% while the intuition model gave the same performance as earlier. Intuition model did not improve in its performance even though it has longer time of execution. Intuition failed to improve as it is a mapping function based on attributes and does not depend on logical thinking or learning. Therefore, intuition process cannot be a replacement for the logic based methods, but they can surely be a necessary addition to the current approaches in the field of problem solving and artificial general intelligence. Please also note that 0% error was never obtained in any of the experiments due to the inclusion of unknown entities in the datasets.

We have noticed that intuition tends to give us unexpected or new answers. In such scenarios, the answer is not based on some random experience result of a problem. However, the method used to obtain the result is the actual experience set element value here and not the result. This is because the intuition process considers the importance of the method at a higher value than the result derived from the method. Also, another scenario in the above case is when the mapping from the experience set element value for another problem set is returned as the actual or correct value for our problem at hand. Such mapping of wrong experience set is taken as correct because the importance and priority values are high for this case. The intuition model considers this value as the correct value and thus is returned back as the value to the user.

Intuition can go wrong because intuition just maps it to the element which falls in the most optimal fit according to the attributes mentioned above. That is the reason why INTUITION CAN ALSO GO WRONG. The answers can go wrong because of the importance of:-
1) The mapping may be done to the incorrect element due to the assignment of incorrect importance levels to the same process. This will give it higher priority than the other element (which is assumed to be correct logically) and thus the incorrect assignment happens.
2) It is also possible that the past experience has become less important in the scenario and some other scenario has taken precedence. Thus the user may just ignore the knowledge set and the problem may cease to exist. This can also lead to mapping to incorrect values.

To conclude, we believe that in intuition, when a problem comes, it is mapped to a knowledge set element (past experience) from the brain. This element will have attributes and values that define its entire structure and function. One of the main points in the above algorithm is the mapping ability of the experience set element to the problem set element. Logical processes calculate the entire process as logical entities. However, the intuition model maps the past experience, processes it with adjustments and then presents it to the user. Several methods such as Bayesian networks, neural networks and Hidden Markov Models consider the use of logical processes in their implementation. The limitation of these methods is the implementation of unknown entities that may add up, or existing entities that may change state or even be removed from the problem scope. They are unable to handle unknown entities [4, 5] as the ability to find a solution to them is absent from the current knowledge using logical approaches. However, the current intuition model here proposes that such entities can be handled due to the mapping of past experience elements in a symbolical or artistic manner. This seemingly non-logical approach gives answers to the unknown entities in the most optimum manner (the result is also adjusted based on current time and other conditions).

We hope that our future work will consider uncertainty better and create a model that can be more reliable and accurate. We also hope that this paper will stimulate more research in mapping higher level human intelligence in AI (artificial intelligence) and machine learning. We also need to look at increasing the accuracy of the intuition model and develop it further for more complex scenarios. Finally, we would like to look at the hypothesis that dreams, imagination and creativity are related in function and structure to our current intuition model and hope to investigate them in future.

**Conflicts of Interests**

The authors cite no conflicts of interests.

**Acknowledgement**

Many thanks to friends and family for helping in completing this research project.

Table -1) Results (Percentage of errors) for the Datasets Evaluation using the Car Evaluation and Poker Evaluation. Note that we have brought in different values for each cycle.

| Car Evaluation Dataset Executed Using Methods Below(rounded off value) | | | | | | |
|---|---|---|---|---|---|---|
| Cycle No | Neural Networks | | Intuition Model | | Hidden Markov Models | |
| | Untrained | Trained | Untrained | Trained | Untrained | Trained |
| 1 | 32% | 3% | 12% | 13% | 22 % | 3% |
| 2 | 30 % | 2% | 17% | 19% | 27% | 5% |
| 3 | 38% | 4% | 13% | 11% | 19% | 2% |
| 4 | 39% | 8% | 17% | 16% | 31% | 3% |
| 5 | 41% | 1% | 19% | 18% | 26% | 1% |
| | | | | | | |

| Poker Evaluation Dataset Executed Using Methods Below(rounded off value) | | | | | | |
|---|---|---|---|---|---|---|
| Cycle No | Neural Networks | | Intuition Model | | Hidden Markov Models | |
| | Untrained | Trained | Untrained | Trained | Untrained | Trained |
| 1 | 33% | 2% | 11% | 12% | 22 % | 7% |
| 2 | 40 % | 1% | 15% | 16% | 33% | 2% |
| 3 | 37% | 6% | 14% | 16% | 16% | 3% |
| 4 | 25% | 4% | 18% | 19% | 36% | 9% |
| 5 | 41% | 5% | 17% | 15% | 29% | 4% |
| | | | | | | |